\documentclass[letterpaper,10pt,conference]{ieeeconf}

\IEEEoverridecommandlockouts
\usepackage[utf8]{inputenc}

\usepackage{refcount}
\usepackage{footmisc}
\usepackage{cite}
\usepackage{xcolor}
\usepackage{amsmath,amssymb}
\usepackage{bm}
\usepackage{algorithm}
\usepackage{algpseudocode}
\usepackage{booktabs}
\usepackage{tikz}
\usepackage{url}
\usepackage{adjustbox}
\usepackage{soul}
\usepackage{multirow}

\usepackage{microtype}
\usepackage{xurl} 
\newtheorem{theorem}{Theorem}

\newtheorem{assumption}{Assumption}

\newtheorem{remark}{Remark}

\newcommand{\Tm}{\mathcal{T}}
\newcommand{\Cm}{\mathcal{C}}
\newcommand{\Sm}{\mathcal{S}}

\newcommand{\R}{\mathbb{R}}

\newcommand{\Kset}{\mathcal{K}}
\newcommand{\Uset}{\mathcal{U}}
\newcommand{\Xset}{\mathcal{X}}
\newcommand{\Zset}{\mathcal{Z}}

\newcommand{\Bset}{\mathcal{B}}
\newcommand{\Ig}{\mathcal{I}}
\newcommand{\Jteam}{\mathcal{J}}

\newcommand{\Hp}{H_{T}}
\newcommand{\Hpr}{H_{0}}

\newcommand{\cen}{\mathrm{cen}}

\newcommand{\avar}{\mathrm{CVaR}}

\newcommand{\norm}[1]{\left\lVert #1 \right\rVert}
\newcommand{\tr}{\operatorname{tr}}

\DeclareMathOperator{\diag}{diag}

\DeclareMathOperator{\grad}{grad}
\newcommand{\Exp}{\mathrm{Exp}}

\newcommand{\wed}{^{\wedge}}

\newcommand{\Sph}{\mathbb{S}}

\newcommand{\nader}[1]{{\color{blue} #1}}

\newcommand{\qiyu}[1]{{\color{brown} #1}}

\begin{document}

\title{\huge \bf TRACE: Privacy-Preserving Next-Best-View Selection over Distributed 3D Gaussian-Splat Maps}

\author{Amirhossein Mollaei Khass, Athanasios Cosse, Qiyu Sun, Nader Motee
\thanks{
This work was supported in part by the ONR under grants numbers  N00014-23-1-2779 and N00014-26-1-2246. \newline
A.M. Khass, A. Cosse, and N. Motee are with the Department of Mechanical Engineering and Mechanics, Lehigh University, Bethlehem, PA, 18015, USA. {\tt\small \{ammb23,asc425,motee\}@lehigh.edu}.\endgraf
}
}

\maketitle

\begin{abstract}
Share the light, not the map. We study next-best-view selection for a team of robots,
each of which builds its own 3D Gaussian Splatting map and keeps it private. A robot
picks the view with the largest expected information gain (EIG) about the splats along
its own path. This gain depends on the other maps. Their splats occlude its own and shine behind them,
so the gain has to be evaluated against the pooled map. No robot has this map. We show that the coupling passes through only two ray quantities, the transmittance
in front of a splat and the radiance behind it, and that both are sums over the hits of
the ray. Hence, they decompose across the robots and each robot sums them over depth bins in its own map, along the rays of a
candidate view, and sends the sums with their pose derivatives. The robot planning the
view turns them into its EIG and gradient on $SO(3)$. Transmittance and Radiance
Aggregates, Communicated for the EIG, give the protocol its name: TRACE. No robot shares its splats, and the message
size does not grow with a map. We prove that the reconstruction is exact unless a depth
bin behind a splat mixes hits of two robots, and we bound the error otherwise. Over 100
next-best-view decisions in Habitat-Sim, TRACE picks a heading within $15^\circ$ of the
centralized one in $83.3\%$ of the cases, and its views reach $97.9\%$ of the
centralized EIG.
\end{abstract}

\section{Introduction}
\label{sec:introduction}
A team of robots enters a building after an earthquake. Each robot maps what it sees and
plans where to look next, in order to reconstruct the part of the scene it must traverse
well enough to move through safely. The difficulty is that no robot sees everything it
needs from where it stands. The far side of a corner ahead of it is already behind a
second robot, and the top of a shelf it passes under was seen by a third that flew over
it. Such viewpoints are out of reach, yet they hold the geometry that keeps the first
robot safe. The value of a view is therefore a team
quantity~\cite{Di_NeRF,MAC_Ego3D,khass2026multi}. A robot that scores its views on its
own map alone looks where the team has already looked, and still misses what it cannot
see. Coordinating these views is hard under limited bandwidth, energy and
computation~\cite{HAMMER,MAGiC_SLAM,Zhao2025RSS}.

We build these maps with 3D Gaussian Splatting (3DGS)~\cite{kerbl20233d}. A 3DGS map is
a set of anisotropic Gaussian primitives, and a view is rendered by projecting them onto
the image plane and compositing them front to back along each
ray~\cite{ye2025gsplat,Matsuki}. The render is differentiable in the parameters of the
map and in the pose of the camera. Next-best-view (NBV) methods exploit this property to
pick poses that reduce the uncertainty of the
map~\cite{Fisherrf,xue2026gavis,khass2025active,chen2026cover}. The rendering Jacobian
gives the sensitivities that build a Fisher information objective, which is then optimized over the pose.

In a team of robots, this objective couples the maps at the level of the render. A
centralized method sends the observations or the local maps to a server, which builds
one 3DGS map and evaluates the team objective on
it~\cite{Chen_Su_Wang_Han_Xia_Bu_Li_Hu_Meng_Wang_2026,li2026compact}. A distributed
method keeps them on board, and then no robot holds the whole scene. Along a shared ray,
the contribution of a primitive depends on the light that reaches it through the hits in
front of it, and on the radiance behind it. Both are set by the joint depth order of all
primitives on the ray, which no robot knows from its own map. A robot that renders only
its own map therefore biases the objective and its gradient.

We propose TRACE, a distributed NBV method that rebuilds these two quantities without
sharing observations or Gaussian parameters. Along the rays of a candidate view, each
robot reports the \textbf{t}ransmittance and \textbf{r}adiance \textbf{a}ggregates
\textbf{c}ommunicated for the \textbf{E}IG, which is where the name comes from. These
aggregates are sums over depth bins. Hence, they carry what the other maps place along a
ray without revealing which splats placed it. Each robot then evaluates the pooled-map
gradient on its own and turns its camera, with no consensus over viewpoints.

\vspace{0.1cm}
\noindent{\bf Main contributions.}
(i) We show that the EIG of a view, and its pose-gradient, depend on the maps of the
other robots only through two quantities on each ray, the transmittance in front of a
splat and the radiance behind it. (ii) We derive the pose-gradient of the masked EIG on
$SO(3)$ in closed form from the outputs of a standard rasterizer. (iii) We propose
TRACE, a protocol in which the robots exchange these two quantities in depth bins, with
their pose derivatives, and never their splats; the message size does not depend on the
number of splats. (iv) We prove that the EIG and its gradient are exact when no depth
bin behind a splat mixes hits of two robots, and we bound the error otherwise. (v) We
test TRACE in simulation against the centralized computation.

\begin{figure*}[t]
\centering
\vspace{1.8pt}
\includegraphics[width=\textwidth]{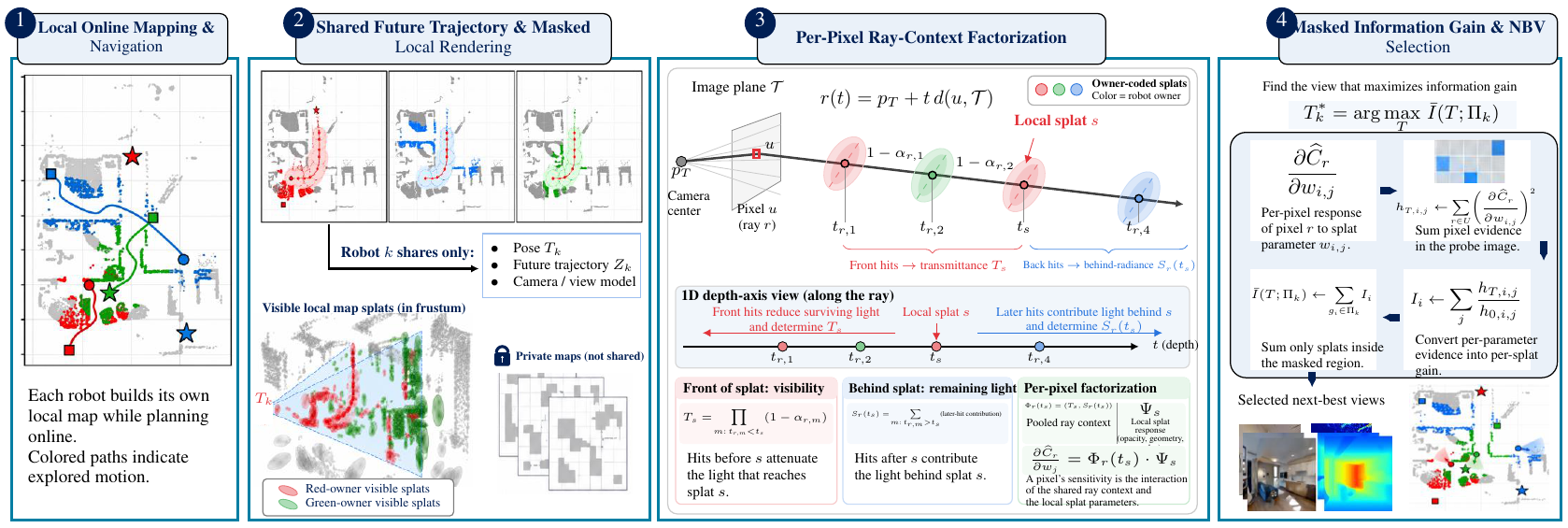}
\caption{Overview of TRACE: robots maintain private local 3DGS maps, construct trajectory-conditioned masks, exchange factorized ray context, and optimize masked information gain for distributed NBV selection.}
\label{fig:system_overview}
\vspace{-0.4cm}
\end{figure*}

\vspace{0.1cm}
\noindent{\bf Related work.}
NBV methods select camera poses that maximize information gain or improve coverage.
FisherRF~\cite{Fisherrf} scores a view by the trace of a diagonal Fisher information
matrix, and COVER~\cite{chen2026cover} builds an efficient coverage surrogate from an
approximation of it. GAVIS~\cite{xue2026gavis} models view-dependent uncertainty with an
anisotropic visibility field, and RT-GuIDE~\cite{RT_guide} uses the updates of the
Gaussian parameters as a cheap information proxy. Others add risk-aware
masks~\cite{liu2024beyond} or couple information gathering with safety-critical
control~\cite{khass2026conflict,khass2026splat}. Across a team, views are classically picked from
finite candidate sets by greedy submodular maximization on one shared
map~\cite{lauri2020multi}. All of these score a view on a map the planner can access.
Multi-agent neural mapping solves the other half of the problem, which is to build one
consistent model from distributed observations. Di-NeRF~\cite{Di_NeRF} trains a NeRF by
consensus and RAMEN~\cite{Zhao2025RSS} maps asynchronously. For 3DGS,
MAGiC-SLAM~\cite{MAGiC_SLAM} merges local submaps by loop closures,
MAC-Ego3D~\cite{MAC_Ego3D} enforces geometric consistency by Gaussian consensus, and
CoMA-SLAM~\cite{Chen_Su_Wang_Han_Xia_Bu_Li_Hu_Meng_Wang_2026} works on Gaussian
surfels, while others cut communication with compact
submaps~\cite{HAMMER,li2026compact}. All of these build one map of the whole scene.
Closest to our goal is distributed risk-averse NBV over 3DGS
maps~\cite{khass2026multi}, where each robot scores its views on its own map and
consensus reconciles the team. That work distributes the optimization, and the splats of the other robots never enter a local render. TRACE distributes the render instead. No global map is formed, and each robot evaluates the gain of the pooled map from what it receives along the rays of its own view.


\section{Preliminaries and Definitions}\label{sec:background}

\noindent{\bf 3D Gaussian Splatting.}
A 3D Gaussian Splatting (3DGS) map is a finite set $\Xset=\{x_i\}$ of anisotropic
Gaussian primitives. Splat $x_i$ has a mean $\mu_i\in\R^3$, a covariance
$\Sigma_i=R(q_i)\diag(\ell_i)^2R(q_i)^\top$ with unit quaternion $q_i$ and scales
$\ell_i\in\R^3_{+}$, a peak opacity $\epsilon_i\in[0,1]$, and a view-dependent color
$c_i\in\R^{C}$ represented by spherical-harmonic coefficients $\theta_i$ of order $L$. We
collect the parameters of $x_i$ in the ordered vector
$w_i=(\mu_i,q_i,\ell_i,\epsilon_i,\theta_i)=(w_{i,1},\dots,w_{i,n_w})^{\!\top}\in\R^{n_w}$, which
lists the three coordinates of the mean, the four components of the quaternion, the
three scales, the opacity, and the $C(L+1)^2$ coefficients of the color. Thus, 
$n_w=11+C(L+1)^2$, where for $L=3$ and $C=3$, $n_w=59$. The index $j$ selects one scalar
parameter $w_{i,j}$ of the splat, and $w\in\R^{|\Xset|n_w}$ is the concatenation of the
$w_i$ over the entire map.

A camera pose, or view, $T\in SE(3)$ consists of a rotation $R_T\in SO(3)$
and a center $p_T\in\R^3$. Pixel $u$, with homogeneous coordinates $\tilde u=(u,1)$,
casts the ray $r(t)=p_T+t\,R_T\check d(u)$, $t\ge0$, where
$\check d(u):=K^{-1}\tilde u/\norm{K^{-1}\tilde u}$ is its camera-frame direction and $K$
is the intrinsic matrix. Along ray $r$ with image coordinate $u_r$, splat $x_i$ has the
blending coefficient~\cite{kerbl20233d,ye2025gsplat}
\begin{equation}\label{eq:blend}
\alpha_{r,i}=\epsilon_i\exp\bigl(-\tfrac12(u_r-\mu'_i)^\top(\Sigma'_i)^{-1}(u_r-\mu'_i)\bigr),
\end{equation}
where $\mu'_i$ and $\Sigma'_i$ are the EWA projections of the camera-frame mean
$\mu_{c,i}=R_T^\top(\mu_i-p_T)$ and covariance $\Sigma_{c,i}=R_T^\top\Sigma_iR_T$. We say
that splat $x_i$ is a hit of ray $r$ if
$(u_r-\mu'_i)^\top(\Sigma'_i)^{-1}(u_r-\mu'_i)\le\varsigma^2$ with $\varsigma=3$. Each
Gaussian is truncated at this level, so every ray has finitely many hits. The
coefficient $\alpha_{r,i}$ is smooth in the splat parameters and in the
pose. We take the depth of a hit to be
the radial distance $t_{r,i}:=\norm{\mu_i-p_T}$ and order the hits front to back,
$t_{r,1}<t_{r,2}<\cdots$; we explain this choice in Section~\ref{sec:grad}.

\vspace{0.3cm}

\noindent{\bf Transmittance and Compositing.}
The fraction of light that survives all hits ahead of depth $t$ is the transmittance
\begin{equation}\label{eq:transdef}
\Tm_r(t)=\!\!\prod_{m:\,t_{r,m}<t}\!\!(1-\alpha_{r,m}),
\vspace{-0.2cm}
\end{equation}
and the pixel color is the front-to-back alpha composite
\begin{equation}\label{eq:comp}
\widehat\Cm_r=\sum_{m}c_m\,\alpha_{r,m}\,\Tm_r(t_{r,m}^-)\in\R^{C},
\vspace{-0.2cm}
\end{equation}
where $t^-_{r,m}$ is the depth just in front of hit $m$ and $c_m$ is its color evaluated
at the direction from $p_T$ to $\mu_m$. Let $\Uset$ be the set of pixels, or rays,
$r=1,\dots,n_u$. Stacking \eqref{eq:comp} over all pixels gives the rendering model
\begin{equation}\label{eq:rendermodel}
f(T,w):=\bigl(\widehat\Cm_1,\dots,\widehat\Cm_{n_u}\bigr)^{\!\top}\in\R^{n_uC}.
\vspace{-0.2cm}
\end{equation}
The model maps a pose and the map parameters to the rendered image, one entry per pixel
and channel~\cite{Fisherrf,liu2024beyond}. The
pose enters $f$ only through the hit sets and the coefficients $\alpha_{r,m}$. The derivative of $f$ in $w$ is the rendering Jacobian $J=\nabla_wf$. The derivative in $T$ gives the pose-gradients of Section~\ref{sec:grad}.

\vspace{0.3cm}
\noindent{\bf Behind-Radiance and Opacity Sensitivity.}
The behind-radiance at depth $t$ is the light composited by the hits behind it,
\begin{equation}\label{eq:behind}
\Sm_r(t):=\!\!\sum_{m:\,t_{r,m}>t}\!\!c_m\,\alpha_{r,m}\,\Tm_r(t_{r,m}^-)\in\R^{C}.
\end{equation}
Differentiating \eqref{eq:comp} with respect to the blending coefficient of hit $s$ gives
the opacity sensitivity \cite{ye2025gsplat}
\begin{equation}\label{eq:tug}
\frac{\partial\widehat\Cm_r}{\partial\alpha_s}=c_s\Tm_s-\frac{\Sm_r(t_s)}{1-\alpha_s},
\qquad \Tm_s:=\Tm_r(t_s^-).
\end{equation}
Hence, the sensitivity of a splat depends on the rest of the scene only through the
occlusion in front of it, $\Tm_s$, and the radiance behind it, $\Sm_r(t_s)$.
Section~\ref{sec:distributed} shows that these two quantities are all that couples the
private maps.

\vspace{0.3cm}
\noindent{\bf Risk-Aware Masked Zone.}
We focus perception where safety is at stake, using the risk field
of~\cite{liu2024beyond,khass2025active,khass2026multi}. For a point $y$, the signed distance
$\mathrm{sd}(y,x_i)$ to splat $x_i\sim\mathcal N(\mu_i,\Sigma_i)$ along a fixed direction
is Gaussian, so its Conditional Value-at-Risk at level $\varepsilon$ has a closed form,
and the point risk is the conservative minimum over splats,
\begin{equation}\label{eq:riskfield}
\varrho_\varepsilon(y):=\min_{x_i\in\Xset}\avar_\varepsilon\bigl(\mathrm{sd}(y,x_i)\bigr).
\end{equation}
Given a reference trajectory $\Zset=\{z_1,\dots,z_{n_z}\}$, each waypoint receives the
risk-shaped radius $r_{\mathrm{mask}}(z_k)=\beta_1e^{-\beta_2\varrho_\varepsilon(z_k)}$,
$\beta_1,\beta_2>0$, larger where the risk is higher, and the masked zone is the union of
balls $\Pi_{\Zset}=\bigcup_k\Bset(z_k,r_{\mathrm{mask}}(z_k))$. It selects the
safety-relevant splats
$$\Xset_{\Pi}:=\{x_i\in\Xset\mid\mu_i\in\Pi_{\Zset}\},\vspace{-0.2cm}$$ over which all information objectives are evaluated.

\section{Expected Information Gain of a View}\label{sec:eig}

The composite \eqref{eq:comp} is nonlinear in the map parameters. Thus, we linearize
the rendering model \eqref{eq:rendermodel} around the current estimate $w_0$,
\begin{equation}\label{eq:L1}
f(T,w)\approx f(T,w_0)+J\,(w-w_0),
\end{equation}
where $J:=\nabla_w f(T,w)\big|_{w_0}\in\R^{n_uC\times n_w|\Xset|}$ is the rendering
Jacobian. Under Gaussian pixel noise with variance $\sigma^2$, the Fisher information
of the view is
\begin{equation}\label{eq:fim}
\Hp=\sigma^{-2}J^\top J\in\R^{n_w|\Xset|\times n_w|\Xset|}.
\end{equation}
It is additive over pixels and over views. When the robot takes the view, the
accumulated prior is updated as $\Hpr\mapsto\Hpr+\Hp$.

We measure the value of a view by the D-optimal gain
$\log\det(\Hpr+\Hp)-\log\det\Hpr$. We use its first-order expansion\footnote{With $M:=\Hpr^{-1}\Hp$, we have
$0\le\tr(M)-[\log\det(\Hpr+\Hp)-\log\det\Hpr]\le\tfrac12\tr(M^2)$. This follows
from $0\le x-\log(1+x)\le x^2/2$ for $x\ge0$.}, which is the
trace $\tr(\Hpr^{-1}\Hp)$. Following FisherRF~\cite{Fisherrf}, we
store the prior in diagonal form,
$[\Hpr]_{ii}=\diag\bigl(h_{0,i1},\dots,h_{0,in_w}\bigr)$.  A full $\Hpr$ over the whole map is too large to store or invert. We write
$J=[J_1,\dots,J_{|\Xset|}]$, where $J_i\in\R^{n_uC\times n_w}$ contains the columns of
splat $x_i$. Since $\Hpr$ is diagonal, we have
$\tr(\Hpr^{-1}\Hp)=\sum_i\tr\bigl([\Hpr]_{ii}^{-1}[\Hp]_{ii}\bigr)$. This sum uses only
the diagonal blocks $[\Hp]_{ii}=\sigma^{-2}J_i^\top J_i$. Thus, the gain is additive over
splats even though $\Hp$ is a full matrix. The $j$-th diagonal entry of $[\Hp]_{ii}$ is
\begin{equation}\label{eq:hdiag}
h_{T,ij}=\sigma^{-2}\sum_{r\in\Uset}\sum_{\mathrm{ch}=1}^{C}
\left(\frac{\partial\widehat\Cm_{r,\mathrm{ch}}}{\partial w_{i,j}}\right)^{\!2}.
\end{equation}
Thus, $h_{T,ij}$ is the information that the view carries about $w_{i,j}$, and
$h_{0,ij}$ is the information already accumulated about it. The information
gain of splat $x_i$ is defined as
\begin{equation}\label{eq:Ii}
\Ig_i:=\tr\bigl([\Hp]_{ii}[\Hpr]_{ii}^{-1}\bigr)=\sum_{j=1}^{n_w}\frac{h_{T,ij}}{h_{0,ij}}.
\end{equation}
The masked expected information gain (EIG) of the view is the sum over the masked
splats,
\begin{equation}\label{eq:eig}
\bar\Ig(T;\Pi)=\sum_{x_i\in\Xset_{\Pi}}\Ig_i .
\end{equation}

\begin{assumption}[Regularity]\label{as:regular}
(i) $\alpha_{r,m}\le\alpha_{\max}<1$ for every ray and hit; (ii) $\Hpr\succeq\lambda I$
for some $\lambda>0$, so that $h_{0,ij}\ge\lambda$; (iii) the pose derivatives are taken
at a pose around which the hit set of every pixel does not change.\footnote{Part (i)
holds because 3DGS implementations clamp the blending coefficient, typically at
$\alpha_{\max}=0.99$. Part (ii) holds because the prior is initialized with the
regularizer $\lambda I$. For part (iii), a splat enters or leaves the hit set of a pixel
only when its boundary crosses the pixel center, and \eqref{eq:comp} jumps there. This
happens on a set of poses of measure zero. Elsewhere, a small rotation keeps all hit
sets fixed and \eqref{eq:comp} is smooth in the pose.}
\end{assumption}

The masked set $\Xset_{\Pi}$ is fixed by the trajectory, and the prior entries
$h_{0,ij}$ are fixed by the past views. Neither depends on the candidate pose. Thus, the
pose enters \eqref{eq:eig} only through the sensitivities
$\partial\widehat\Cm_r/\partial w_{i,j}$ in \eqref{eq:hdiag}. Our results also hold for
the log-determinant gain $\sum_j\log(1+h_{T,ij}/h_{0,ij})$, since it depends on the same
entries $h_{T,ij}$. The only change is the weight $1/h_{0,ij}$ in \eqref{eq:so3chain},
which becomes $1/(h_{0,ij}+h_{T,ij})$. When the splat is clear from the context, we drop the first index to simplify the notation and write $w_j$, $h_{T,j}$, and $h_{0,j}$.

\section{Problem Statement}\label{sec:problem}

Let $\Kset=\{1,\dots,N\}$ index the robots. Robot $k$ owns a private map $\Xset_k$ with
parameters $w_k:=(w_i)_{x_i\in\Xset_k}$. Privacy means that robot $k$ never discloses
the parameters of its splats, or how many it has. The pooled map of the team is
$\Xset=\bigcup_{l\in\Kset}\Xset_l$ with parameters $w=(w_1,\dots,w_N)$. Robot $k$ moves
along its own trajectory $\Zset_k$ with the masked zone $\Pi_k:=\Pi_{\Zset_k}$. Its
safety-relevant splats come from the pooled map,
$\Xset_{\Pi_k}=\{x_i\in\Xset\mid\mu_i\in\Pi_k\}$, and we write
$\Xset_{l,\Pi_k}:=\Xset_l\cap\Xset_{\Pi_k}$ for those owned by robot $l$.

Let us consider a candidate view $T_k$ of robot $k$. The render \eqref{eq:comp} of the pooled
map from $T_k$ uses the hits of all robots along each ray, with the joint transmittance
$\Tm_r$ of \eqref{eq:transdef}. Robot $k$ alone can render only its own map. This
render uses its own hits $\alpha^{(k)}_{r,m}$ and the local transmittance
$\Tm^{(k)}_r$, which is \eqref{eq:transdef} over the hits of robot $k$ only. The two
renders differ, because the splats of the other robots occlude the splats of robot $k$
and shine behind them through $\Tm_s$ and $\Sm_r(t_s)$ in \eqref{eq:tug}.

We apply \eqref{eq:eig} to the pooled map and the zone $\Pi_k$. This gives the
centralized EIG of the view,
\begin{equation}\label{eq:cenEIG}
\begin{aligned}
\bar\Ig^{\cen}(T_k;\Pi_k)&:=\sum_{x_i\in\Xset_{\Pi_k}}\Ig_i^{\cen}(T_k)
=\sum_{l\in\Kset}\Ig_{l\to k}(T_k),\\
\Ig_{l\to k}(T_k)&:=\sum_{x_i\in\Xset_{l,\Pi_k}}\Ig_i^{\cen}(T_k),
\end{aligned}
\end{equation}
where $\Ig_i^{\cen}$ is the gain \eqref{eq:Ii} of splat $x_i$ computed with the render
of the pooled map, and $\Ig_{l\to k}$ is the contribution of the splats of robot $l$ to
the view of robot $k$. The team selects its next best views by solving
\begin{equation}\label{cent-NBV}
\hspace{-0.2cm}\underset{T_1,\dots,T_N\in SE(3)}{\mathrm{maximize}}\quad
\Jteam(T_1,\dots,T_N):=\sum_{k\in\Kset}\bar\Ig^{\cen}(T_k;\Pi_k).
\end{equation}
The pose $T_k$ appears only in the $k$-th summand. Thus, the decision variables are
decoupled, and each robot can ascend the gradient of its own $\bar\Ig^{\cen}(T_k;\Pi_k)$.
No consensus among the views is needed \cite{Zhao2025RSS, khass2026multi}. The robots are coupled only through the pooled
map, against which every summand is evaluated. Robot $k$ does not have this map. It has
only its own splats, and privacy prevents the other robots from sharing theirs. The
problem is to compute $\bar\Ig^{\cen}(T_k;\Pi_k)$ and its pose-gradient at every robot
$k$ while preserving the privacy of all maps. We solve it with a privacy-preserving
scheme in which the robots exchange only what the render of the pooled map needs along
the rays of the candidate view, and never the splats themselves. For simplicity of
notation, we write $\bar\Ig^{\cen}(T_k)$ for $\bar\Ig^{\cen}(T_k;\Pi_k)$.

\begin{assumption}\label{distinct-disjoint}
(i) No two hits share a depth on any ray. (ii) The maps are disjoint,
$\Xset_k\cap\Xset_l=\emptyset$ for $k\ne l$, so that $w=(w_1,\dots,w_N)$ lists every
splat once.\footnote{Both conditions hold in practice. Splat means are estimated from
noisy images, so two hits at exactly the same depth on a ray is an event of probability
zero. Each robot builds its own splats from its own observations. This does not mean
that the maps do not spatially overlap. It only means that two robots observing the same scene do
not create two identical splats.}
\end{assumption}

\section{Pose-Gradient of the EIG on $SO(3)$}\label{sec:grad}

We fix a robot and drop its index $k$. The trajectory fixes its position, so the
optimization variable in \eqref{cent-NBV} is the rotation $R\in SO(3)$. We parameterize
it about the current orientation $R_0$ as $R(\xi)=R_0\Exp(\xi)$, where $\Exp(\xi)=e^{\xi\wed}$
and $(\cdot)\wed:\R^3\to\mathfrak{so}(3)$ is the hat operator, $\xi\wed v=\xi\times v$.
Thus, $\xi\in\R^3$ is a small rotation of the camera about its own axes, and $\xi=0$ is the
current orientation. When the camera rotates, the world direction of every pixel
changes. This changes the blending coefficients $\alpha_{r,m}$, and with them the
transmittance $\Tm_r(t;\xi)$ and the rendered color $\widehat\Cm_r(\xi)$, while the splat
parameters stay at $w_0$. The gradient of $\bar\Ig^{\cen}$ in this parameterization
is\footnote{All pose derivatives are taken in the parameterization $R=R_0\Exp(\xi)$ and
evaluated at $\xi=0$: for a generator index $a\in\{1,2,3\}$,
$\partial\varphi/\partial\xi_a$ denotes
$\partial\varphi(R_0\Exp(\xi))/\partial\xi_a|_{\xi=0}$, so that
$\grad\varphi(R_0)=\sum_{a=1}^{3}(\partial\varphi/\partial\xi_a)\,e_a$ in the body frame
and the directional derivative along $\xi$ is
$D_{\xi}\varphi:=\grad\varphi(R_0)\cdot\xi$.}
\begin{equation}\label{eq:so3chain}
\hspace{-0.3cm} \bigl[\grad\bar\Ig^{\cen}(R_0)\bigr]_a=
\!\!\sum_{x_i\in\Xset_{\Pi}}\ \sum_{j=1}^{n_w}\ \sum_{r\in\Uset}
\zeta_{r,ij}^{\top}\,
\frac{\partial}{\partial \xi_a}
\!\left(\frac{\partial\hspace{0.04cm}\widehat\Cm_{r}}{\partial w_{i,j}}\right),
\end{equation}
where $a=1,2,3$ and
$\zeta_{r,ij}:=\frac{2}{\sigma^{2}h_{0,ij}}
\frac{\partial\widehat\Cm_{r}}{\partial w_{i,j}}\in\R^{C}$. The ascent step is
$R^{+}=R_0\Exp\bigl(\eta\,\grad\bar\Ig^{\cen}\bigr)$ with $\eta>0$, and $R_0$ is reset
to $R^{+}$ after each step. The first factor in \eqref{eq:so3chain} is the pixel
sensitivity returned by the backward pass of the rasterizer, and the second factor is
its derivative in the pose. We now make this pose dependence explicit.

Since $p_T$ is fixed, the pose enters the blending coefficient \eqref{eq:blend} only
through the camera-frame mean and covariance $\mu_{c,m}(\xi)=R(\xi)^\top(\mu_m-p_T)$ and
$\Sigma_{c,m}(\xi)=R(\xi)^\top\Sigma_mR(\xi)$. By differentiating them at $\xi=0$, we get 
\begin{equation}\label{mu-Sigma-derivatives}
\frac{\partial\mu_{c,m}}{\partial \xi_a}=\mu_{c,m}\times e_a,~~
\frac{\partial\Sigma_{c,m}}{\partial \xi_a}=\Sigma_{c,m}\,e_a\wed-e_a\wed\,\Sigma_{c,m},
\end{equation}
where $\{e_1,e_2,e_3\}$ is the standard basis of $\R^3$. By the chain rule, we have 
\begin{equation}\label{eq:dalpha}
\frac{\partial\alpha_{r,m}}{\partial \xi_a}
=\left(\frac{\partial\alpha_{r,m}}{\partial\mu_{c,m}}\right)^{\!\top}
\frac{\partial\mu_{c,m}}{\partial \xi_a}
+\Bigl\langle\frac{\partial\alpha_{r,m}}{\partial\Sigma_{c,m}},\;
\frac{\partial\Sigma_{c,m}}{\partial \xi_a}\Bigr\rangle_{\!F},
\end{equation}
with $\langle A,B\rangle_F=\tr(A^\top B)$. The two partial derivatives in $\mu_{c,m}$
and $\Sigma_{c,m}$ are those of the EWA projection at a fixed pixel, which every
differentiable rasterizer forms in its backward pass~\cite{kerbl20233d,ye2025gsplat}. If the rasterizer does not expose them, $\partial\alpha_{r,m}/\partial\xi_a$ follows
from central differences of the robot's own render at $R_0\Exp(\pm he_a)$. In either
case, the owner of the splat computes $\partial\alpha_{r,m}/\partial\xi_a$ for each
pixel and hit from its own map.

We use two conventions in the rest of the paper. First, the depth of a hit is the radial
distance $t_{r,m}=\norm{\mu_m-p_T}$, and the depth grid of Section~\ref{sec:distributed}
is a grid in radial depth. This distance does not depend on $R$ or on the pixel. Thus,
the depth, the front-to-back order and the bin of a hit do not change with $\xi$, while
the view-space depth $e_3^\top\mu_{c,m}$ does. Second, the color $c_m$ of a hit is
evaluated at the direction from $p_T$ to $\mu_m$, which a rotation about $p_T$ does not
move. Thus, $c_m$ does not change with $\xi$ either. Since the hit sets are fixed by Assumption~\ref{as:regular}(iii), the pose acts on
\eqref{eq:comp} only through the blending coefficients. Hence, all pose derivatives below are obtained from $\partial\alpha_{r,m}/\partial\xi_a$
by the chain rule.

The rasterizer of a robot returns $\partial\widehat\Cm_r/\partial w_{i,j}$ for its own
splats. This is the first factor in \eqref{eq:so3chain}. The difficulty is that this
factor and its derivative in $\xi$ must be evaluated on the pooled scene, not on the
robot's own map. By \eqref{eq:tug}, the pooled scene enters through only two ray
quantities: the transmittance $\Tm_s$ in front of the splat and the behind-radiance
$\Sm_r(t_s)$. On a ray that meets another map, the robot cannot compute either one from
its own splats. Section~\ref{sec:distributed} shows how the robots obtain them without
sharing their splats.

\section{TRACE: Distributed Computation of the Pose-Gradient}\label{sec:distributed}

Suppose that a robot renders only its own map. Then, it misses the splats of the other
maps. On every ray that meets another map, it overestimates the transmittance $\Tm_s$
and computes the wrong behind-radiance $\Sm_r(t_s)$. This biases both the EIG and its
gradient \eqref{eq:so3chain}. TRACE removes this bias. Each robot computes two depth
profiles from its own map along the rays that see the masked zone, together with their
pose derivatives, and shares them with the other robots. Due to the page limit, we only sketch the proofs.

\subsection{Cross-map coupling}\label{sub:factor}
Let us fix a ray $r$, and let $s$ be the hit of the splat we differentiate. We drop the
splat index, as in Section~\ref{sec:eig}, and write $\alpha_s$, $c_s$ and $t_s$ for its
blending coefficient, color and radial depth, and $w_j$ for one entry of its parameter
vector. Let $g_j:=\partial\alpha_s/\partial w_j$ and $\eta_j:=\partial c_s/\partial w_j$.
The owner of the splat computes both from its own map. We have $g_j=0$ for the harmonic
coefficients, and $\eta_j=0$ for the opacity, the quaternion and the scales.

By Assumption~\ref{as:regular}(iii), the parameter $w_j$ enters \eqref{eq:comp} only
through $c_s$ and $\alpha_s$. We have $\partial\widehat\Cm_r/\partial c_s=\alpha_s\Tm_s$,
and $\partial\widehat\Cm_r/\partial\alpha_s$ is given by \eqref{eq:tug}. Thus, the chain
rule gives
\begin{equation}\label{eq:factor}
\frac{\partial\widehat\Cm_r}{\partial w_j}
=\Tm_s\bigl(\alpha_s\eta_j+c_sg_j\bigr)-\frac{\Sm_r(t_s)}{1-\alpha_s}\,g_j
=:\Phi_r(t_s)\cdot\Psi_s,
\end{equation}
where $\Phi_r(t_s):=\bigl(\Tm_s,\Sm_r(t_s)\bigr)$ and
$\Psi_s:=\bigl(\alpha_s\eta_j+c_sg_j,\;-g_j/(1-\alpha_s)\bigr)$, and the product of two
such pairs is $(\Tm,\Sm)\cdot(\Psi_1,\Psi_2):=\Tm\Psi_1+\Sm\Psi_2\in\R^{C}$. The pair
$\Phi_r(t_s)$ is the context of the hit. It says how much light reaches the hit and what
lies behind it, and it depends on every map. The pair $\Psi_s$ is the local factor. It
depends on the differentiated splat alone. The next theorem shows that the pose
derivative decomposes along the same two pairs.

\begin{theorem}\label{thm:sep}
Let Assumption~\ref{as:regular} hold. For every ray $r$, hit $s$, parameter $w_j$ and
$a\in\{1,2,3\}$,
\begin{equation}\label{eq:splitterm}
\frac{\partial}{\partial\xi_a}\!\left(\frac{\partial\widehat\Cm_r}{\partial w_j}\right)
=\underbrace{\frac{\partial\Phi_r(t_s)}{\partial\xi_a}\cdot\Psi_s}_{\text{context}}
\;+\;\underbrace{\Phi_r(t_s)\cdot\frac{\partial\Psi_s}{\partial\xi_a}}_{\text{local}},
\end{equation}
where
\begin{equation*}\label{eq:Dpsi}
\frac{\partial\Psi_s}{\partial\xi_a}
=\Bigl(\eta_j,\;-\frac{g_j}{(1-\alpha_s)^2}\Bigr)\frac{\partial\alpha_s}{\partial\xi_a}
+\Bigl(c_s,\;-\frac{1}{1-\alpha_s}\Bigr)\frac{\partial^2\alpha_s}{\partial\xi_a\,\partial w_j}.
\end{equation*}
\end{theorem}
Equation \eqref{eq:splitterm} is the product rule applied to \eqref{eq:factor}. Equation
for $\partial\Psi_s/\partial\xi_a$ holds because $c_s$ and $\eta_j$ do not change with $\xi$, by
Section~\ref{sec:grad}. Thus, $\Psi_s$ depends on the pose only through $\alpha_s$ and
$g_j$. The owner computes $\Psi_s$ and $\partial\Psi_s/\partial\xi_a$ from its own map.
It takes $\partial\alpha_s/\partial\xi_a$ from \eqref{eq:dalpha}, and it obtains the
mixed derivative $\partial^2\alpha_s/\partial\xi_a\partial w_j$ by central differences of
$g_j$. The mixed derivative vanishes only for the harmonic coefficients, where $g_j=0$.
Hence, we cannot drop it.

What the owner cannot compute on a ray that meets another map is the context
$\Phi_r(t_s)$ and its three pose derivatives. These are $4(1+C)$ numbers per masked hit.
By Section~\ref{sec:problem}, $\bar\Ig^{\cen}=\sum_{l\in\Kset}\Ig_{l\to k}$, so the
gradient \eqref{eq:so3chain} splits into the blocks
$G_{l\to k}:=\grad\Ig_{l\to k}(R_0)\in\R^{3}$. By \eqref{eq:factor} and
\eqref{eq:splitterm}, each block depends on the other maps only through the context.
Hence, robot $l$ computes $G_{l\to k}$ from its own map once it is given these numbers,
and nothing else about the other maps crosses between robots.

\subsection{Reconstructing the context from depth profiles}\label{sub:profiles}
We fix the pose at which the gradient is evaluated, and a grid $t_0<t_1<\cdots<t_B$ in
radial depth with bin $b:=(t_{b-1},t_b]$ of width $\delta$. Along pixel ray $r$, robot
$l$ computes from its own map the two profiles
\begin{equation}\label{eq:profiles}
\resizebox{0.89\columnwidth}{!}{$\displaystyle
\hspace{-0.2cm}\tau^{(l)}_r[b]:=\sum_{m\in b}-\log\bigl(1-\alpha_{r,m}\bigr),~~
\gamma^{(l)}_r[b]:=\sum_{m\in b}c_m\alpha_{r,m}\Tm^{(l)}_r(t_m^-)
$}\vspace{-0.15cm}
\end{equation}
where $m\in b$ ranges over the hits of map $l$ in bin $b$, and $\Tm^{(l)}_r$ is the
transmittance of map $l$ alone. The first profile is the optical depth that map $l$ places in the bin. The second is the radiance it composites there against its own transmittance. Both are sums over the hits
of a bin, so the robots exchange aggregate light along the rays, and never the
parameters, or the number, of their splats. These sums are many to one, so the splats
cannot be recovered from them. What they disclose is a coarse density and color volume
of each map, at the resolution of a pixel in angle and $\delta$ in depth, inside the
footprint of the receiver's masked zone.

In the same pass, robot $l$ differentiates both profiles term by term with
\eqref{eq:dalpha}. At each of its own masked hits $s$ in bin $b$, it also records two
exact own factors, its transmittance $\Tm^{(l)}_r(t_s^-)$ in front of $s$ and the
radiance $\widetilde\Gamma^{(l)}_{>s}:=\sum_{m\in b,\,t_m>t_s}c_m\alpha_{r,m}\Tm^{(l)}_r(t_m^-)$
of its own hits behind $s$, with their pose derivatives.

We now consider a masked hit $s$ of robot $k$ on ray $r$, in bin $b$; we drop the pixel
subscript to simplify the notation. The profiles are bin sums, so they do not say where a hit sits inside its bin. This matters, because the hits of bin $b$ in front of $s$ belong to $\Tm_s$ and
those behind it to $\Sm_r(t_s)$. Hence, the behind-radiance splits into three parts,
\begin{equation}\label{eq:Ssplit}
\hspace{-0.2cm}\Sm_r(t_s)=\underbrace{\Sm_r(t_b)}_{\substack{\text{hits in the}\\ \text{bins behind }b}}
\;+\;\underbrace{\Gamma^{(k)}_{>s}}_{\substack{\text{robot }k\text{'s hits in}\\ \text{bin }b\text{ behind }s}}
\;+\;\underbrace{\Gamma^{(-k)}_{>s}}_{\substack{\text{the other robots' hits}\\ \text{in bin }b\text{ behind }s}}.
\end{equation}
This identity is exact, and we recover its three parts from the messages. The maps are
disjoint by Assumption~\ref{distinct-disjoint}, so the pooled transmittance factors by
owner, and the receiver knows the factor of map $l$ at the bin boundaries,
$\hat\Tm^{(l)}[b]:=\exp(-\sum_{b'\le b}\tau^{(l)}[b'])$. It therefore attenuates the
radiance of each bin by the other maps' transmittance at the end of that bin,
$\Gamma^{(l)}[b']:=\gamma^{(l)}[b']\prod_{l'\ne l}\hat\Tm^{(l')}[b']$, and attenuates
$\widetilde\Gamma^{(k)}_{>s}$ in the same way. Only the share of the other robots in bin $b$ cannot be located inside the bin.
Therefore, it is interpolated linearly, with $\theta_s:=(t_s-t_{b-1})/\delta\in(0,1]$ the
position of $s$ in its bin. Applying this to
\eqref{eq:Ssplit}, and to the transmittance in front of $s$, gives
\begin{gather}
\resizebox{0.88\columnwidth}{!}{$\displaystyle
\widehat\Sm(t_s)=\sum_{b'>b}\sum_{l}\Gamma^{(l)}[b']
+\prod_{l\ne k}\hat\Tm^{(l)}[b]\,\widetilde\Gamma^{(k)}_{>s}
+(1-\theta_s)\sum_{l\ne k}\Gamma^{(l)}[b]$},\label{eq:Shat}\\
\resizebox{0.86\columnwidth}{!}{$\displaystyle
\log\widehat\Tm_s=\log\Tm^{(k)}(t_s^-)
+\sum_{l\ne k}\bigl[\log\hat\Tm^{(l)}[b-1]-\theta_s\tau^{(l)}[b]\bigr]$}.\label{eq:That}
\vspace{-0.2cm}
\end{gather}
The pose derivatives of $\widehat\Phi_r(t_s):=(\widehat\Tm_s,\widehat\Sm(t_s))$ follow by
differentiating \eqref{eq:Shat} and \eqref{eq:That} term by term, with the received
derivative profiles and the receiver's own derivatives of $\Tm^{(k)}(t_s^-)$ and
$\widetilde\Gamma^{(k)}_{>s}$. We say bin $b'$ of ray $r$ mixed if it holds hits of two
distinct robots.

\begin{theorem}[Exact reconstruction]
\label{thm:recon}
Suppose that Assumptions \ref{as:regular} and \ref{distinct-disjoint} hold and that all
robots compute the profiles \eqref{eq:profiles} at the same pose and on the same depth
grid. Consider a hit $s$ of ray $r$ that belongs to robot $k$ and lies in bin $b$, with
context $\Phi_r(t_s)$ in the pooled map and estimate $\widehat\Phi_r(t_s)$ from
\eqref{eq:Shat}--\eqref{eq:That}. If every bin $b'\ge b$ of ray $r$ holds hits of at most
one robot, then
\[
\widehat\Phi_r(t_s)=\Phi_r(t_s)
~~\text{and}~~
\frac{\partial\widehat\Phi_r(t_s)}{\partial\xi_a}=\frac{\partial\Phi_r(t_s)}{\partial\xi_a}~~\textrm{for}~~ a=1,2,3.
\]
\end{theorem}
Bins in front of $b$ may be mixed, since only boundary values of the transmittance enter
there and these factor by owner exactly. When a bin at or behind $s$ is mixed, the
receiver bounds its error from the received optical depths. Let us write
$\Delta_{-l}[b']:=\sum_{l'\ne l}\tau^{(l')}[b']$ for the optical depth that the other
robots place in bin $b'$. For nonnegative colors,
$|\log\widehat\Tm_s-\log\Tm_s|\le\Delta_{-k}[b]$, each radiance term of \eqref{eq:Shat}
is off by at most the factor $e^{\Delta_{-l}[b']}-1$ of its own size, and the
interpolated share of bin $b$ by at most $e^{\Delta_{-l}[b]}$ times its own size. Both
bounds vanish under the condition of Theorem~\ref{thm:recon}. Two robots mix a bin only
when their hits on the same ray lie within $\delta$ of each other. Thus, the bin width
trades message size against a certified error.

\subsection{The TRACE protocol}\label{sub:protocol}
We replace the exact context $\Phi_r(t_s)$ by its estimate $\widehat\Phi_r(t_s)$ in
\eqref{eq:factor} and \eqref{eq:splitterm}. Then \eqref{eq:hdiag}, \eqref{eq:Ii} and
\eqref{eq:so3chain} give the contribution of a masked splat $x_i$ to the EIG of the view
and to its gradient,
\begin{equation*}
\begin{aligned}
\widehat\Ig^{\cen}_i&=\sum_{j=1}^{n_w}\frac{1}{\sigma^{2}h_{0,ij}}\sum_{r\in\Uset}
\bigl\|\widehat\Phi_r(t_s)\cdot\Psi_s\bigr\|^{2},\\
\bigl[\widehat{\grad}\,\Ig^{\cen}_i\bigr]_a&=
\sum_{j=1}^{n_w}\sum_{r\in\Uset}\widehat\zeta_{r,ij}^{\top}
\Bigl[\frac{\partial\widehat\Phi_r(t_s)}{\partial\xi_a}\cdot\Psi_s
+\widehat\Phi_r(t_s)\cdot\frac{\partial\Psi_s}{\partial\xi_a}\Bigr],
\end{aligned}
\vspace{-0.2cm}
\end{equation*}
where $s$ is the hit of $x_i$ on ray $r$ and
$\widehat\zeta_{r,ij}:=\frac{2}{\sigma^{2}h_{0,ij}}\widehat\Phi_r(t_s)\cdot\Psi_s$.
Robot $l$ evaluates these two sums for its own masked splats in robot $k$'s zone. It adds
them into $\widehat\Ig_{l\to k}$ and $\widehat G_{l\to k}$, its estimates of
$\Ig_{l\to k}$ and $G_{l\to k}$. The EIG of robot $k$'s view and its gradient are the
sums of these estimates over $l$. Robot $k$ then takes the ascent step
$R^{+}=R_0\Exp\bigl(\eta\sum_{l}\widehat G_{l\to k}\bigr)$. By Theorem~\ref{thm:recon},
if no bin at or behind a masked hit is mixed on any ray, then
$\widehat\Ig^{\cen}_i=\Ig^{\cen}_i$ for every $x_i\in\Xset_{\Pi_k}$ and
$\widehat{\grad}\,\bar\Ig^{\cen}(R_0)=\grad\bar\Ig^{\cen}(R_0)$.

The whole cycle is distributed. No robot ever forms the pooled map, and there is no
central node. One cycle at robot $k$ runs as follows. Robot $k$ announces its center,
camera model, depth grid and masked zone once per cycle, and its orientation once per
ascent step. At every step, each robot renders its own map from robot $k$'s pose on the
footprint, the pixels whose rays meet the zone, and broadcasts its two profiles with
their derivatives; the bins in front of the zone hold no masked hit, so they are summed
into one optical depth per pixel. Each robot then fuses the received profiles with its
own factors, evaluates its own masked hits, and returns the four numbers
$(\widehat\Ig_{l\to k},\widehat G_{l\to k})$ to robot $k$. All robots run this cycle for
their own views at the same time, since the views are decoupled in \eqref{cent-NBV}. The
message of a robot per step is $n^{\Pi}_u[4(1+C)\bar B+4]$ scalars, with $n^{\Pi}_u$
footprint pixels and $\bar B$ occupied bins per pixel, and it does not depend on the
number of splats. Thus, the computation and the communication scale with the footprint of
the masked zone, and not with the size of the maps.

TRACE computes a centralized quantity in a distributed way. It is exact when no bin at or
behind a masked hit mixes two robots, and the received optical depths bound the error
otherwise.

\begin{table*}[t]
\centering
\vspace{1.8mm}
\setlength{\tabcolsep}{3pt}
\renewcommand{\arraystretch}{1.05}
\scriptsize
\begin{tabular}{@{}l@{\hskip 8pt}ccc@{\hskip 5pt}ccc@{\hskip 9pt}ccc@{\hskip 5pt}ccc@{\hskip 9pt}ccc@{}}
\toprule
\multirow{3}{*}{Method}
& \multicolumn{6}{c}{Agents $N=3$ (Cantwell)} & \multicolumn{6}{c}{Agents $N=7$ (Frankton)} & \multicolumn{3}{c}{Agents $N=10$ (Frankton)} \\
\cmidrule(lr){2-7}\cmidrule(lr){8-13}\cmidrule(l){14-16}
& \multicolumn{3}{c}{Navigation} & \multicolumn{3}{c}{Exploration}
& \multicolumn{3}{c}{Navigation} & \multicolumn{3}{c}{Exploration}
& \multicolumn{3}{c}{Exploration} \\
\cmidrule(lr){2-4}\cmidrule(lr){5-7}\cmidrule(lr){8-10}\cmidrule(lr){11-13}\cmidrule(l){14-16}
& PSNR $\uparrow$ & SSIM$\uparrow$ & LPIPS$\downarrow$ & PSNR$\uparrow$ & SSIM$\uparrow$ & LPIPS$\downarrow$
& PSNR$\uparrow$ & SSIM$\uparrow$ & LPIPS $\downarrow$ & PSNR$\uparrow$ & SSIM$\uparrow$ & LPIPS$\downarrow$
& PSNR$\uparrow$ & SSIM$\uparrow$ & LPIPS $\downarrow$ \\
\midrule
Decentralized
& 20.40 & .647 & .468 &  {19.00} &  {.601} &  {.402}
&   {20.66} &   {.659} &   {.502} &  {18.53} &   {.501} &  {.475}
&   {18.50} &   {.553} &  {.478}
\\
Centralized Map-Sharing
&  {21.19} &  {.697} &  {.459} & 18.93 & .599 & .410
&  {22.63} &  {.762} &  {.347} & 18.69 & .565 &   {.471}
&  {20.35} &  {.699} &   {.433} \\
Centralized Oracle
&   {24.55} &   {.877} &   {.290} &   {21.22} &   {.785} &   {.368}
& 24.13 & .832 & .252 & 21.86 & .776 & .371
& 21.11 & .708 & .343 \\
\midrule
TRACE (ours)
&  {21.07} &  {.653} &  {.467} &  {19.64} &  {.603} &  {.393}
& 21.74 & .705 & .370 & 18.84 & .569 & .440
& {20.01} & {.621} & {.45} \\
TRACE-Sparse (ours)
& 19.88 & .610 & .497 & 19.30 & .601 & .402
& 21.07 & .673 & .379 &   {18.64} &  {.512} & .460
& 19.24 & .603 & .477 \\
\bottomrule
\end{tabular}
\caption{Reconstruction quality across five Methods on two tasks, and team sizes $N\!\in\!\{3,7,10\}$. All metrics are evaluated at the poses each run actually executed and therefore
describe the map along the trajectory each planner produced.}
\label{tab:recon}
\vspace{-8.5pt}
\vspace{-0.25cm}
\end{table*}


\begin{table}[t]
\centering
\vspace{-2pt}
\scriptsize
\setlength{\tabcolsep}{1.7pt}
\renewcommand{\arraystretch}{0.92}
\resizebox{\columnwidth}{!}{%
\begin{tabular}{@{}lrrrrrrrrr@{}}
\toprule
\multirow{2}{*}{Method}
& \multicolumn{3}{c}{Comm. (MB/step)}
& \multicolumn{3}{c}{NBV time (s)}
& \multicolumn{3}{c}{Map size ($10^3$)} \\
\cmidrule(lr){2-4}
\cmidrule(lr){5-7}
\cmidrule(l){8-10}
& $N=3$ & $N=7$ & $N=10$
& $N=3$ & $N=7$ & $N=10$
& $N=3$ & $N=7$ & $N=10$ \\
\midrule
Map Sharing
& 18.73 & 67.24 & 89.05
& 4.97 & 5.37 & 6.79
& 138.2 & 213.3 & 188.3 \\

Oracle
& {0.27} & {0.87} & {1.14}
& 3.82 & 5.82 & 6.34
& 175.5 & 248.0 & 248.0 \\
\midrule
TRACE
& 1004.48 & 1547.89 & 2055.25
& 2.91 & 3.04 & 3.51
& 135.4 & 138.0 & 218.4 \\

TRACE-Sparse
& 1.45 & 4.91 & 8.42
& {1.21} & {1.46} & {2.01}
& 126.4 & 209.2 & 203.2 \\
\bottomrule
\end{tabular}%
}
\caption{Communication, NBV time, and mean map size during
multi-robot exploration.}
\label{tab:efficiency}
\vspace{-4pt}
\end{table}

\section{Experiments}
\label{sec:experiments}

We evaluate our proposed distributed framework, TRACE, against decentralized and centralized baselines with $N\in\{3,4,7,10\}$ robots. We assess reconstruction quality, communication, NBV computation, and map size, followed by an ablation of the TRACE communication variants.
The multi-robot 3DGS systems discussed in Section~\ref{sec:introduction} primarily address map construction, alignment, or fusion, whereas TRACE addresses NBV selection over private local maps. An end-to-end comparison with those systems would mix differences in localization and mapping with the effect of NBV selection and conflate the view-selection policy. We therefore hold the simulator, incremental mapper, planning pipeline, and evaluation protocol fixed, and compare the selection policy. 

\subsection{Setup and baselines}
\label{subsec}

Experiments use Habitat-Sim~\cite{habitat19iccv} and Gibson indoor scenes~\cite{xia2018gibson}. Each robot carries a $400\times400$ RGB-D camera with a $90^\circ$ horizontal field of view. Except for the Centralized Oracle, robot $k$ maintains a private incremental 3DGS map $\mathcal{G}_k$ implemented with gsplat~\cite{ye2025gsplat}. At each planning cycle, the robot updates its occupancy map, assigns a task-dependent goal or frontier, computes an $A^\star$ path with $0.3,\mathrm{m}$ waypoint spacing, and constructs $\Pi_k$ over the next five waypoints using $\beta_1=2.0$, $\beta_2=3.0$, and $\varepsilon=0.1$. It then selects an NBV, advances along the path, and incorporates the new observation into its map. An episode terminates upon task completion or reaching the prescribed step limit. All methods share the scene, seed, start/goal, camera, and step limit. Experiments run on an NVIDIA RTX A2000 GPU.

The decentralized baseline requires no communication. {Centralized map-sharing} transmits local splat updates to a server that maintains a fused global map. {Centralized oracle} streams observations to a server that maintains the single team map and returns each NBV. Our {TRACE} method keeps observations and Gaussian parameters local and exchanges dense ray--depth profiles with their derivatives to recover the global map EIG gradient. We additionally evaluate {TRACE-Sparse}, a communication-oriented variant that transmits only quantized occupied profiles and yaw derivatives.
\vspace{-0.1cm}
\subsection{Reconstruction quality}
\label{subsec:reconstruction}
Table~\ref{tab:recon} evaluates the reconstruction produced by each NBV strategy. After each run, the resulting map is rendered from camera poses around the executed trajectory, and the metrics are averaged over all evaluated poses and robots. TRACE consistently improves reconstruction over Decentralized planning across team sizes. Because the robots retain private maps and exchange neither raw observations nor Gaussian parameters, these gains show that the proposed ray profiles provide useful cross-map context for selecting complementary views that cannot be identified from a local map alone. TRACE also remains competitive with Centralized Map Sharing, outperforming it in some settings while remaining close in others. TRACE recovers much of the benefit of global map planning without fusing independently optimized and uncertain local 3DGS maps, which can introduce inconsistencies and reconstruction artifacts. Centralized Oracle achieves the highest quality because all observations are integrated into a single jointly optimized team map and therefore serves as the centralized reference. Finally, TRACE consistently outperforms TRACE-Sparse in PSNR, showing that profile sparsification and quantization introduce an accuracy--communication trade-off.

\begin{figure}[t]
    \centering
    \includegraphics[
width=0.62\textwidth,
        trim={7.1cm 5.2cm 5.3cm 4.7cm},
        clip
    ]{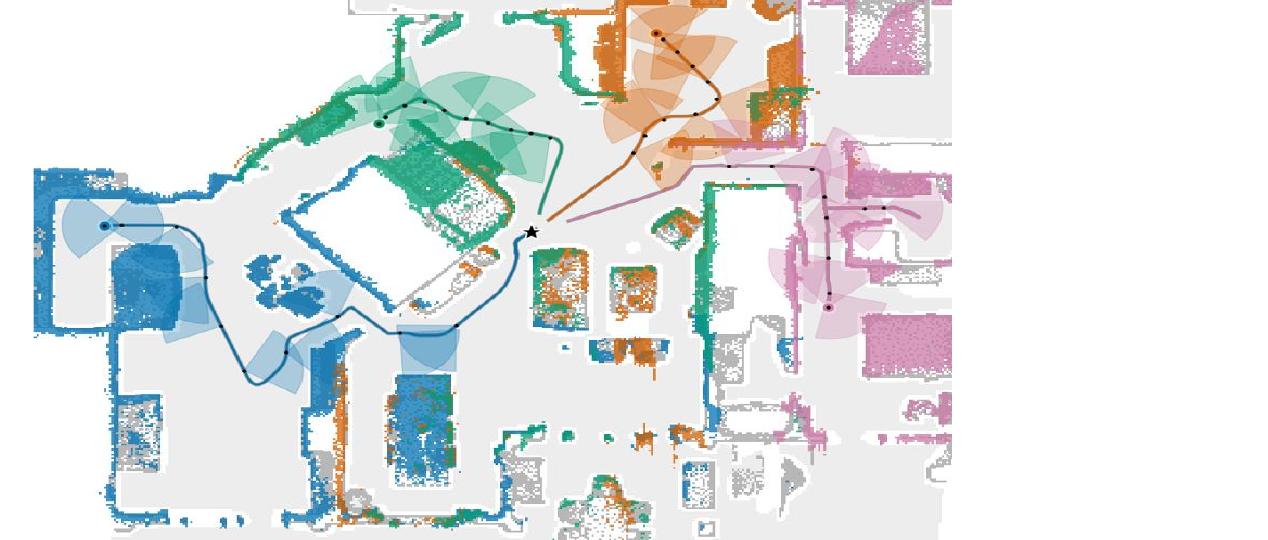}
\caption{Four robots navigating toward a goal in the Lakeville scene. Each color represents a robot’s private local 3DGS map and trajectory.
}
\vspace{-10pt}
\label{fig:trajview}
\vspace{-0.3cm}
\end{figure}

\subsection{Communication, computation, and map scaling}
\label{subsec:efficiency}

Table~\ref{tab:efficiency} reports communication and computation during exploration. Communication is the average team payload per planning step. NBV time is the mean view-selection time over all decisions, including profile construction, rendering, and optimization but excluding network delay. Map size is the final number of Gaussians stored by each robot, averaged across the team; Centralized Oracle instead maintains one shared map.
Centralized Map Sharing transmits local Gaussian updates, so its communication increases with the number of robots and their map sizes. Centralized Oracle sends RGB-D observations and poses, producing a smaller payload; however, its smaller payload does not imply map privacy. TRACE exchanges optical-depth and behind-radiance profiles with their derivatives over the shared rays and depth bins; its payload therefore depends on map overlap and team size. TRACE-Sparse transmits only occupied entries on a coarser depth grid and retains derivatives along the optimized yaw axis. These reductions lower communication by approximately three orders of magnitude relative to TRACE and by over $90\%$ relative to Centralized Map Sharing.

Both TRACE variants keep RGB-D observations and Gaussian parameters onboard and exchange only aggregated ray-level statistics. TRACE remains faster than both centralized methods because NBV computation is distributed across the robots, although the reported time excludes network transfer. Centralized Oracle is slowest because all candidates are evaluated against the complete shared map. TRACE-Sparse is fastest because its coarse, occupied-only profiles reduce profile processing and serialization.

\subsection{Ablation study}
\label{subsec:ablation}
\begin{figure}[t]
    \centering
    \includegraphics[
width=0.69\textwidth,
        trim={5.9cm 6.4cm 2.5cm 4.0cm},
        clip
    ]{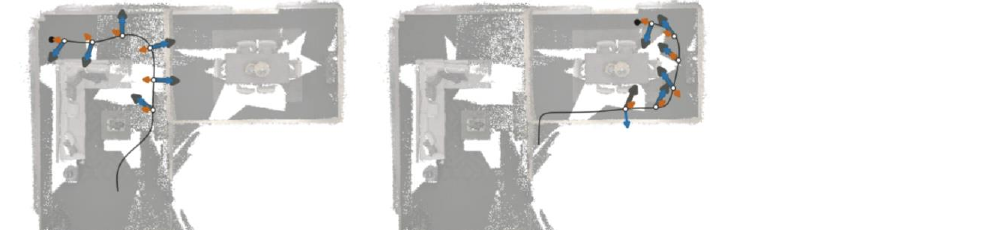}
\caption{Representative NBV selections along two agents' trajectories:
\textcolor{gray}{\rule{2mm}{2mm}} centralized,
\textcolor{blue}{\rule{2mm}{2mm}} TRACE, and
\textcolor{orange}{\rule{2mm}{2mm}} decentralized headings
at each waypoint.
}
\label{fig:exp2explain}
\vspace{-0.5cm}
\end{figure}
Table~\ref{tab:protocol_ablation} compares two map-private TRACE variants. TRACE-Sparse selects each view using the communicated EIG gradient, whereas TRACE-Discrete transmits occupied value profiles without derivatives and explicitly evaluates five fixed yaw candidates. Although TRACE-Discrete requires less communication, its multi-candidate evaluation substantially increases NBV time; TRACE-Sparse reduces the average time from $9.46$ to $0.59\,\mathrm{s}$ for $N=2$ and from $22.34$ to $1.69\,\mathrm{s}$ for $N=5$. TRACE-Discrete achieves moderately better reconstruction, consistent with its ability to evaluate larger viewpoint changes, while both variants keep observations and Gaussian parameters onboard and complete all navigation tasks. These results expose the trade-off among communication, computation, and reconstruction fidelity.

\subsection{Centralized-NBV recovery}
\label{subsec:nbv_recovery}

We evaluate whether TRACE recovers the NBV selected using all private maps. Four robots follow partially overlapping trajectories while maintaining independent 3DGS maps. Across three seeds, we evaluate 100 NBV decisions. Each planner optimizes the heading continuously using its available EIG gradient; the resulting orientation is then projected onto the nearest of 24 executable headings, corresponding to a $15^\circ$ actuation resolution. As shown in Table~\ref{tab:lakeville_results}, TRACE exactly matches the centralized heading in $77.8\%$ of decisions, compared with $25.9\%$ for Decentralized, and reaches $83.3\%$ agreement within $15^\circ$. When evaluated using the centralized objective, TRACE retains $97.9\%$ of the optimal centralized EIG. It also reduces inter-robot map overlap, the region observed by multiple robots, from $25.7\%$ to $15.9\%$. Fig.~\ref{fig:exp2explain} further shows that TRACE selections closely follow the centralized headings along representative trajectories.


\begin{table}[t]
\centering
\vspace{1.8mm}
\setlength{\tabcolsep}{2.8pt}
\renewcommand{\arraystretch}{1.08}
\resizebox{\columnwidth}{!}{%
\begin{tabular}{clccccc}
\toprule
& & & &
\multicolumn{3}{c}{Map Reconstruction} \\
\cmidrule(lr){5-7}
Agents & Method &
Comm./Step (MB)$\downarrow$ &
Avg. NBV Time (s)$\downarrow$ &
PSNR$\uparrow$ & SSIM$\uparrow$ & LPIPS$\downarrow$ \\
\midrule
\multirow{2}{*}{2}
& TRACE-Sparse
& 0.35 & 0.59 & 17.99 & 0.635 & 0.423 \\
& TRACE-Discrete
& 0.09 & 9.46 & 18.64 & {0.679} & {0.386} \\
\midrule
\multirow{2}{*}{5}
& TRACE-Sparse
& 3.86 & {1.69} & 17.33 & 0.642 & 0.415 \\
& TRACE-Discrete
& {0.80} & 22.34 & {18.15} & {0.661} & {0.399} \\
\bottomrule
\end{tabular}%
}
\caption{Comparison of sparse gradient based and discrete NBV selection methods.}
\label{tab:protocol_ablation}
\vspace{-0.2cm}
\end{table}

\begin{table}[t]
\centering
\vspace{-5.5pt}
\scriptsize
\setlength{\tabcolsep}{1.50pt}
\renewcommand{\arraystretch}{0.65}
\begin{tabular*}{\columnwidth}{
@{\extracolsep{\fill}}lcccc@{}}
\toprule
Method
& \multicolumn{2}{c}{NBV match $\uparrow$}
& EIG retained $\uparrow$
& Overlap $\downarrow$
\tabularnewline
\cmidrule(lr){2-3}
& Exact
& Within $15^\circ$
& 
&
\tabularnewline
\midrule
Centralized
& 100.0 & 100.0 & 100.0 & 18.0
\tabularnewline
\addlinespace[1pt]
\textbf{TRACE (ours)}
& \textbf{77.8}
& \textbf{83.3}
& \textbf{97.9}
& \textbf{15.9}
\tabularnewline
Decentralized
& 25.9 & 38.9 & 69.2 & 25.7
\tabularnewline
Random
& 4.2 & 12.5 & 46.3 & --
\tabularnewline
\bottomrule
\end{tabular*}
\caption{Agreement with the centralized NBV, retained EIG, and
inter-robot map overlap. All entries are percentages.}
\label{tab:lakeville_results}
\vspace{-1cm}
\end{table}
\section{Conclusion}
\label{sec:conclusion}
We studied next-best-view selection for a team of robots that keep their 3D Gaussian
Splatting maps private. Each robot ascends the masked information gain of its own view
over the camera orientation. This objective is evaluated on the pooled map, and no robot
holds this map. We showed that the pooled map enters the pixel sensitivity, and its pose
derivative, only through the transmittance in front of a splat and the radiance behind
it. Both are sums over the hits of a ray. Thus, they decompose across the robots. This
led to TRACE. Each robot sends the optical depth and the radiance that its own map
places in each depth bin, with their pose derivatives. The receiver then rebuilds the
information gain of its view and its gradient. The reconstruction is exact unless a
depth bin at or behind a hit mixes two robots. Hence, the bin width sets both the
message size and the accuracy of the gradient. The splats are never shared.

{
\footnotesize
\bibliographystyle{IEEEtran}
\bibliography{references}

}

\end{document}